\documentclass[runningheads]{llncs}

\usepackage{eccv}

\usepackage{eccvabbrv}

\usepackage{graphicx}
\usepackage{booktabs}
\usepackage{enumitem}
\usepackage{wrapfig}
\usepackage{colortbl}
\usepackage{xcolor}
\definecolor{rankone}{RGB}{255,120,120}   
\definecolor{ranktwo}{RGB}{255,180,120}   
\definecolor{rankthree}{RGB}{255,220,140} 
\usepackage{multirow}
\usepackage{pifont}
\newcommand{\xmark}{\ding{55}}%

\usepackage[accsupp]{axessibility}  

\usepackage{hyperref}

\usepackage{orcidlink}

\begin{document}

\title{FrameCrafter: Novel View Synthesis \\ as Video Completion} 

\titlerunning{FrameCrafter}

\author{Qi Wu \and
Khiem Vuong \and
Minsik Jeon \and Srinivasa Narasimhan \and Deva Ramanan }

\authorrunning{Q.~Wu et al.}

\newcommand{\MYhref}[3][0000AA]{\href{#2}{\color[HTML]{#1}{\textbf{#3}}}}%

\institute{Carnegie Mellon University, USA\\
{\normalsize \MYhref{https://frame-crafter.github.io}{\texttt{https://frame-crafter.github.io}}}
}

\maketitle

\begin{figure}
    \centering
    \vspace{-18pt}
    \includegraphics[width=\linewidth]{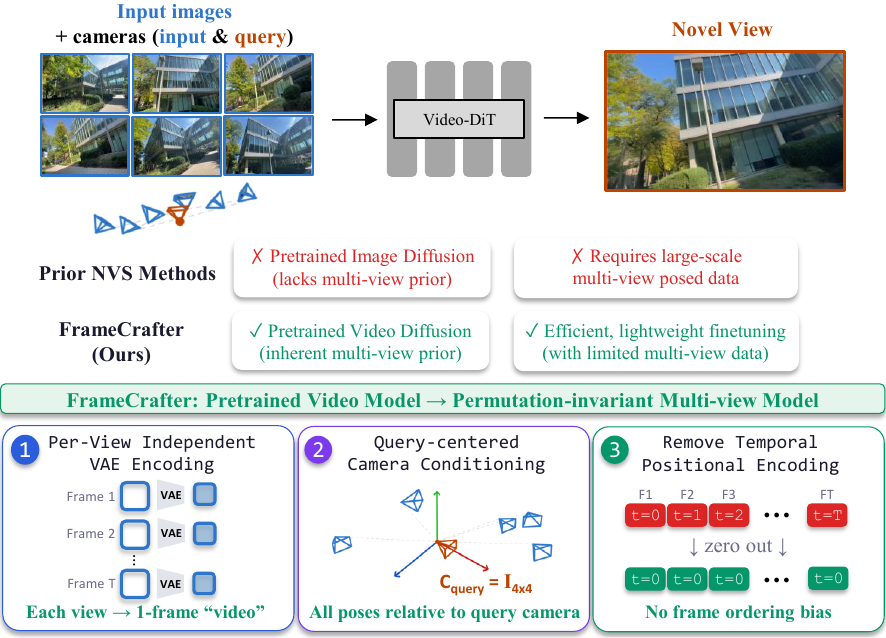}
    \vspace{-8pt}
    \caption{\textbf{Video Diffusion as Multi-view Prior.}
    While prior generative NVS methods typically initialize from image diffusion models and rely on large pose-annotated multi-view datasets, videos naturally capture viewpoint changes and cross-view consistency, making them a more \emph{scalable} source of supervision for NVS. 
    We present \textbf{FrameCrafter}, which adapts pretrained video diffusion models to permutation-invariant multi-view model with lightweight modifications trained using LoRA on only 1,000 DL3DV-10K scenes, achieving strong performance and demonstrating a more data-efficient path to high-fidelity novel view synthesis.}
    \label{fig:teaser}
\end{figure}
\vspace{-0.4in}

\begin{abstract}
  We tackle the problem of sparse novel view synthesis (NVS) using video diffusion models; given $K$ ($\approx 5$) posed images of a scene, we predict the view from a target camera pose. Many prior approaches leverage 
  generative image priors encoded via diffusion models. However, models trained on single images lack multi-view knowledge, and
  we instead argue that video models contain implicit multi-view knowledge and are therefore easier to adapt for NVS. Our key insight is to formulate sparse NVS as a low-frame-rate video completion task. 
  However, one challenge is that sparse NVS is defined over an unordered set of input images, often too sparse to admit a meaningful order, so the models should be {\em invariant} to permutations of that input image set.
  To this end, we present \emph{\textbf{FrameCrafter}}, which adapts video models (naturally trained with coherent frame orderings) to permutation-invariant NVS through simple but effective architectural modifications, including per-frame latent encoding, query-centered camera conditioning and removal of temporal positional embeddings. 
  Our results suggest that video models can be easily modified to ``forget'' about time with minimal supervision, producing state-of-the-art performance on sparse-view NVS benchmarks.
\end{abstract}

\section{Introduction}
\label{sec:intro}

Video diffusion models have advanced rapidly in recent years~\cite{brooks2024video, wan2025, yang2024cogvideox, kong2024hunyuanvideo,blattmann2023stable,chen2023videocrafter1}, demonstrating remarkable capabilities in high-fidelity video generation and strong spatio-temporal consistency. 
Beyond visual realism, these models exhibit increasingly sophisticated world understanding, suggesting an implicit awareness of 3D structure and viewpoint transformation~\cite{huang2025much}. 
Motivated by this geometric consistency, prior works have attempted to repurpose video diffusion models for 3D/4D reconstruction~\cite{huang2025much,jiang2025geo4d} and view-conditioned generation~\cite{chen2025reconstruct, ren2025gen3c, bai2025recammaster, zhu2025aether, xiao2025video4spatial}. 
However, their potential for sparse-view novel view synthesis (NVS) --- particularly under large viewpoint changes --- remains largely underexplored.

Sparse-view novel view synthesis requires a model to understand multi-view relationships and infer geometry that is only partially-observed. 
Regression methods~\cite{mildenhall2021nerf, kerbl20233d} learn deterministic mappings under geometric constraints from input views, but often struggle when geometry is sparsely observed or unseen. 
Image diffusion~\cite{liu2023zero, gao2024cat3d, zhou2025stable, kong2024eschernet}, in contrast, benefits from large-scale generative pretraining and is better equipped to synthesize plausible content under large viewpoint changes.
However, videos, as another scalable source to train diffusion models, are inherently more aligned with the NVS objective: they naturally capture multi-view consistency and viewpoint transitions over time. 
Despite this, most prior diffusion-based methods for sparse-view NVS rely on image diffusion backbones and require substantial multi-view supervision to induce 3D awareness.

Our key thesis is as follows: {\em video} diffusion models provide a stronger prior for sparse-view NVS than {\em image} diffusion models.
We leverage geometric reasoning that has already emerged from large-scale video pretraining, instead of learning multi-view geometry ``from scratch'' on multi-view training data.
While perhaps obvious in retrospect, our approach is nontrivial to operationalize because one must modify a video model to accept unordered collections of views (and cameras) as inputs.

To do so, we introduce \textbf{FrameCrafter} with three key architectural modifications. The first and most important is to modify the video VAE to encode a $K$-frame video as a collection of $K$ independent 1-frame videos, allowing multi-view inputs to be treated as an unordered set.
This is actually a small architectural change, as current image-to-video diffusion models already treat the input image as a 1-frame video during encoding. Secondly, many NVS solutions break permutation invariance by defining camera poses in a world coordinate system aligned to the first input image. Instead, we define world coordinates aligned to the novel query view (to be synthesized). In practice, we find that video diffusion models can be easily taught to condition on cameras by channel-concatenating latent (1-frame) video encodings with Pl\"ucker raymaps. 
Thirdly, we remove the temporal component of 3D Rotary Positional Embeddings (RoPE)~\cite{su2024roformer} so that the model cannot rely on frame indices when processing input views. We demonstrate that our lightweight architectural modifications can turn state-of-the-art video models into state-of-the-art multi-view models that are invariant to permutations of their inputs.
 
Our results show that, with only a small number of (LoRA~\cite{hu2022lora}) trainable parameters and minimal training data (1K scenes from DL3DV~\cite{ling2024dl3dv}), we can finetune video diffusion models to outperform prior image diffusion-based models. 
We further observe a consistent scaling trend across backbones, suggesting that progress in large-scale video foundation models can be directly transferred to NVS without requiring massive curated multi-view datasets.
To our knowledge, FrameCrafter is the first approach to turn video diffusion models into permutation-invariant multi-view generators, leveraging their strong multi-view priors for sparse-view NVS while removing their reliance on temporal ordering.

\section{Related Work}
\subsubsection{Novel View Synthesis.}
Early NVS methods are primarily regression-based, learning radiance fields or renderable scene representations from multi-view inputs using frameworks such as NeRF~\cite{mildenhall2021nerf} and its variants~\cite{yu2021pixelnerf,wang2021ibrnet,chen2021mvsnerf,barron2022mip}, as well as explicit representations such as 3D Gaussian Splatting~\cite{kerbl20233d}. Subsequent works improve robustness via regularization~\cite{niemeyer2022regnerf} or feed-forward scene prediction~\cite{zhang2024gs,jin2024lvsm,jiang2025rayzer,zhao2025rayzer}.
More recently, diffusion-based approaches have emerged as a powerful alternative by leveraging the geometric and appearance priors of pretrained diffusion backbones~\cite{liu2023zero,shi2023zero123++,liu2023syncdreamer,kong2024eschernet,gao2024cat3d,wu2024reconfusion,zhou2025stable}. By incorporating large-scale generative priors, these methods synthesize more realistic novel views and generalize better under large viewpoint changes. Nevertheless, achieving reliable multi-view consistency often relies on substantial pose-annotated multi-view datasets~\cite{zhou2018stereo,reizenstein2021common,ling2024dl3dv,deitke2023objaverse}. This reliance becomes particularly important in sparse-view settings, where limited observations increase geometric ambiguity and place greater demands on cross-view reasoning. Recent state-of-the-art models such as SEVA~\cite{zhou2025stable} demonstrate strong cross-domain performance by combining large pretrained image diffusion backbones with extensive multi-view supervision.
In contrast, we explore video diffusion models as scalable priors for sparse-view NVS, leveraging cross-frame consistency from web-scale video pretraining and directly repurposing pretrained video backbones for geometric reasoning without relying on large curated multi-view datasets.

\subsubsection{Video Diffusion Models and 3D-Aware Video Generation.}
Modern video diffusion architectures~\cite{wan2025, yang2024cogvideox} typically adopt latent diffusion frameworks~\cite{rombach2022high} with spatio-temporal VAEs~\cite{brooks2024video} and transformer-based denoisers~\cite{peebles2023scalable}, achieving strong temporal consistency and scalability through web-scale video pretraining.
Beyond generic video generation, several works incorporate camera control or 3D-aware conditioning into video models. One line of research~\cite{yu2024viewcrafter, ren2025gen3c, chen2025reconstruct, zhi2025mv} first reconstructs a scene and then applies video diffusion to inpaint missing regions during rendering. These approaches provide precise camera trajectory control but rely on off-the-shelf reconstruction modules and pre-rendered intermediate representations.
Another line of work directly conditions video diffusion models on camera poses by injecting pose parameters into the model’s latent representations~\cite{zhu2025aether, xiao2025video4spatial, van2026anyview, wang2024motionctrl} or integrating them into attention mechanisms~\cite{he2024cameractrl, zheng2024cami2v}. These approaches enable camera-controllable generation and demonstrate strong cross-frame coherence without explicit reconstruction. 
However, despite their advances in controllable video synthesis and ability to generate viewpoint changes, the output of such models remains a continuous, temporally ordered video sequence rather than discrete novel view predictions derived from sparse observations. The use of video diffusion models for sparse-view NVS—where both the inputs and outputs are discrete views rather than temporal sequences—remains largely unexplored. 
In this work, we directly repurpose pretrained video diffusion models for sparse-view NVS without relying on pre-reconstruction or predefined camera trajectories. By removing temporal compression and enforcing permutation-invariant per-view encoding, we transform a video generation model into an effective multi-view synthesis backbone with minimal additional supervision.

\section{Method}
\label{sec:method}
\begin{figure}[t]
    \centering
    \includegraphics[width=\linewidth]{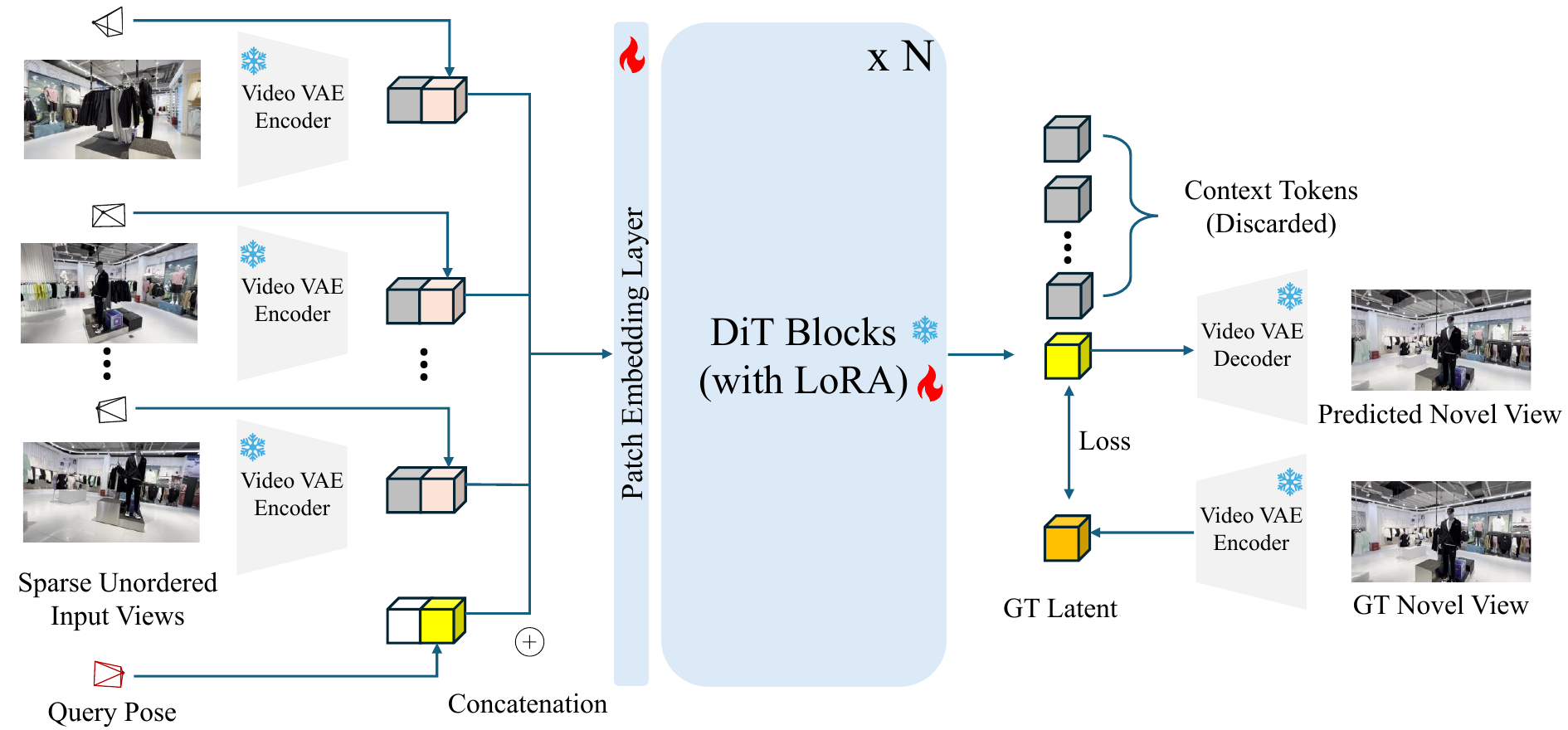}
    \caption{Each input image is encoded by a frozen video VAE, and its camera pose is converted into a Plücker ray map that is concatenated with the image latent along the channel dimension. The resulting view tokens are then concatenated along the temporal dimension, where the first K tokens correspond to input views and the final token represents the query view. Only the patch embedding layer and the LoRA modules in the DiT backbone are trained. During both training and inference, only the predicted query latent is used for decoding and supervision.}
    \vspace{-15pt}
    \label{fig:method}
\end{figure}

We present FrameCrafter, a method for sparse-view novel view synthesis (NVS) by repurposing a pretrained video diffusion model as a multi-view generator.
Given $K$ posed input images of a scene, our model synthesizes a novel target view at a specified camera pose.
The key insight is that a video diffusion transformer, trained on large-scale video data, already possesses strong priors for 3D-consistent scene understanding. We unlock this capability for NVS by (i)~encoding each input view \emph{independently} through the VAE to ensure permutation invariance (Sec.~\ref{sec:separated_encoding}), (ii)~injecting camera geometry via Pl\"ucker ray conditioning while removing the temporal component of 3D Rotary Positional Embeddings to avoid reliance on frame order (Sec.~\ref{sec:camera_conditioning}), and (iii)~fine-tuning with LoRA on multi-view data (Sec.~\ref{sec:training}). An overview of FrameCrafter is shown in Fig.~\ref{fig:method}.

\subsection{Preliminaries}
\label{sec:prelim}

\paragraph{Novel View Synthesis.}
Given a set of $K$ input images $\{\mathbf{I}_k\}_{k=1}^{K}$ of a scene captured from known camera poses $\{\boldsymbol{\pi}_k\}_{k=1}^{K} \in \text{SE}(3)$, novel view synthesis (NVS) aims to generate a photorealistic image $\hat{\mathbf{I}}_{\text{tgt}}$ at a novel target pose $\boldsymbol{\pi}_{\text{tgt}}$.
Classical methods~\cite{mildenhall2021nerf,kerbl20233d} reconstruct explicit or implicit 3D representations before rendering novel views, but degrade with sparse inputs due to limited geometric constraints.
Recent methods~\cite{zhou2025stable,gao2024cat3d,liu2023zero} instead directly predict target views from input images, bypassing 3D reconstruction and relying more on learned priors.

\paragraph{Diffusion Models for Video Generation.}
Diffusion models~\cite{ho2020denoising,song2020score} generate data by learning to reverse a gradual noising process.
In video generation, a latent diffusion framework~\cite{rombach2022high} is commonly adopted: a variational autoencoder (VAE) maps video frames into a compact latent space, and a denoising network--typically a Diffusion Transformer (DiT)~\cite{peebles2023scalable}--operates in this latent domain.

Formally, a video $\mathbf{V} \in \mathbb{R}^{3 \times T \times H \times W}$ is encoded by a VAE encoder $\mathcal{E}$ into latent representations: 
$
    \mathbf{z}_0 = \mathcal{E}(\mathbf{V}) \in \mathbb{R}^{d_z \times T' \times h \times w},
$
where $d_z$ is the latent channel dimension, $h = H / f_s$, $w = W / f_s$ denote spatially downsampled resolutions with factor $f_s$, and $T'$ is the temporally compressed frame count.

During training, noise $\boldsymbol{\epsilon} \sim \mathcal{N}(\mathbf{0}, \mathbf{I})$ is added using a flow-matching schedule~\cite{lipman2022flow}: 
$
    \mathbf{z}_t = (1 - t)\,\mathbf{z}_0 + t\,\boldsymbol{\epsilon}, 
    \text{where }t \in [0, 1],
$
and the DiT $\boldsymbol{\epsilon}_\theta$ is trained to predict the velocity field $\mathbf{v}_t = \boldsymbol{\epsilon} - \mathbf{z}_0$.

\paragraph{Causal 3D VAE in Modern Video Diffusion.}
Modern video diffusion models such as CogVideoX~\cite{yang2024cogvideox} and Wan~\cite{wan2025} employ temporally-causal 3D convolutions in the VAE to ensure future frames do not influence past frames.
The first frame is processed independently as a single-frame ``video'', while each subsequent group of four frames contributes one additional latent timestep, resulting in an effective temporal length of $T' = 1+ (T-1)/4$.
Due to the temporal receptive field of causal convolutions, each frame encoding depends on all preceding frames, introducing a dependency on frame order.
While desirable for video generation, this is undesirable for processing unordered sparse-view inputs.

\subsection{Per-View VAE Encoding for Sparse Views}
\label{sec:separated_encoding}

A naive approach to multi-view NVS with a video diffusion model would concatenate the $K$ input views into a video sequence and encode them jointly through the temporal VAE.
However, this is problematic for sparse-view NVS:  

(1)~Temporal compression degrades per-view information.
The causal video VAE reduces temporal resolution via strided 3D convolutions (typically by a factor of $\sim$4), mapping multiple frames to fewer latent timesteps.
For a video, this may be desirable---it removes redundant information while capturing motion dynamics.
For sparse multi-view images that depict the scene from arbitrarily-ordered viewpoints, this compression is harmful: potentially far away viewpoints can be merged into a shared encoding, losing fine-grained per-view information.

\begin{figure}[t]
    \centering
    \includegraphics[width=\linewidth]{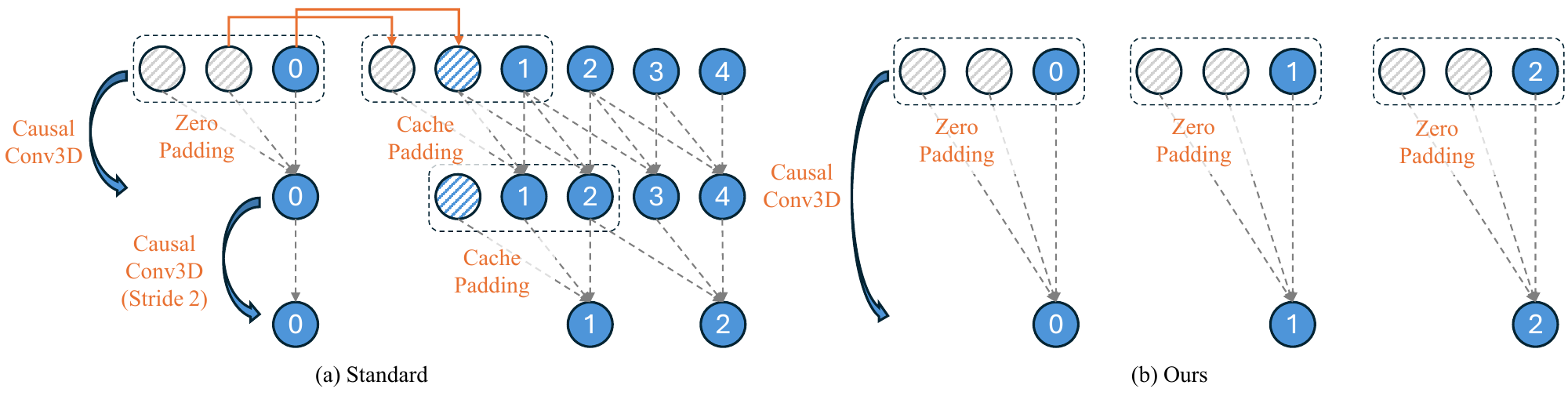}
    \caption{\textbf{Comparison between standard causal video VAE encoding and our per-view independent encoding.}
    (a) Standard causal encoding mixes frame information through feature caching and temporal stride, leading to inter-frame dependency.
    (b) Our per-view encoding eliminates temporal interaction, ensuring independent view-to-latent mapping suitable for unordered sparse-view inputs.}
    \vspace{-15pt}
    \label{fig:vae_compare}
\end{figure}

(2)~Causal ordering violates permutation invariance.
Sparse input views have no natural temporal ordering---the set $\{\mathbf{I}_1, \ldots, \mathbf{I}_K\}$ should be treated as unordered.
The causal 3D convolutions, however, encode a strict left-to-right temporal dependency: the latent of frame $k$ is conditioned on frames $1, \ldots, k-1$ through feature caching.
This means permuting the input order produces different latent representations, introducing an undesirable inductive bias. As illustrated in Fig.~\ref{fig:vae_compare}(a), standard causal VAE encoding propagates multi-frame information through feature caching and temporal downsampling, resulting in ordering dependency and latent compression.

To address these issues, we adopt \emph{per-view independent encoding} (Fig.~\ref{fig:vae_compare}(b)).
Notably, in image-to-video (I2V) generation, the first frame is processed independently without temporal compression, serving as a clean conditioning signal.
Inspired by this design, we similarly encode each input view individually as a single-frame ``video'':
\begin{equation}
    \mathbf{z}_k = \mathcal{E}(\mathbf{I}_k) \in \mathbb{R}^{d_z \times 1 \times h \times w}, \quad k = 1, \ldots, K,
    \label{eq:sep_enc}
\end{equation}
and concatenate the results along the temporal axis:
\begin{equation}
    \mathbf{z}_{\text{ctx}} = [\mathbf{z}_1; \mathbf{z}_2; \ldots; \mathbf{z}_K] \in \mathbb{R}^{d_z \times K \times h \times w}.
    \label{eq:concat}
\end{equation}
For the target view, we encode a zero placeholder $\mathbf{z}_{\text{tgt}} = \mathcal{E}(\mathbf{0})$ and append it:
\begin{equation}
    \mathbf{z}_{\text{all}} = [\mathbf{z}_{\text{ctx}};\; \mathbf{z}_{\text{tgt}}] \in \mathbb{R}^{d_z \times (K+1) \times h \times w}.
    \label{eq:all_latents}
\end{equation}
Since each frame is encoded as a standalone single-frame input, the causal temporal convolutions have no preceding frames to condition on---each view's latent is computed identically and independently of its position.
This guarantees:
\begin{itemize} [nosep,leftmargin=*]
    \item \textbf{Permutation-invariant encoding}: $\mathcal{E}(\mathbf{I}_k)$ is independent of the ordering of the other views.
    \item \textbf{No temporal compression}: each view maps to exactly one latent frame ($T' = 1$ per view), preserving full spatial detail and every view receives the same single-frame encoding treatment, regardless of $K$.
\end{itemize}

During decoding, we similarly decode each latent frame independently:
\begin{equation}
    \hat{\mathbf{I}}_k = \mathcal{D}(\mathbf{z}_k), \quad k = 1, \ldots, K+1,
\end{equation}
where $\mathcal{D}$ is the VAE decoder applied per-frame. This avoids temporal blending artifacts in the decoded output.

\subsection{Camera-Conditioned Diffusion Transformer}
\label{sec:camera_conditioning}

To enable geometric control, we condition the DiT on camera pose information via Pl\"ucker ray maps~\cite{plucker1865xvii,zhang2024cameras}.

\paragraph{Pl\"ucker ray maps.}
For each camera with intrinsic matrix $\mathbf{K}_k$ and camera-to-world extrinsic $\mathbf{T}_k \in \text{SE}(3)$, we compute a dense ray map.
For each pixel $(u, v)$, the ray origin $\mathbf{o}$ and direction $\mathbf{d}$ are computed via back-projection, and the Pl\"ucker representation is:
$
    \mathbf{p}(u,v) = \left[\hat{\mathbf{d}};\; \mathbf{o} \times \hat{\mathbf{d}}\right] \in \mathbb{R}^{6},
    \label{eq:plucker}
$
where $\hat{\mathbf{d}} = \mathbf{d} / \|\mathbf{d}\|$ is the normalized ray direction.
Per-pixel Pl\"ucker coordinates are downsampled to the latent resolution via \emph{pixel unshuffle}~\cite{shi2016real} with factor $f_s$, converting 6 channels at resolution $H{\times}W$ into $6f_s^2$ channels at resolution $h{\times}w$. 
Unlike interpolation-based downsampling, pixel unshuffle preserves per-pixel ray information by rearranging spatial details into channels without loss, maintaining precise geometric correspondence with VAE latents and retaining fine-grained camera geometry.
Much prior work defined cameras (and rays) with respect to a world coordinate system aligned to the first input view~\cite{wang2024dust3r,wang2025vggt}, but this introduces a dependence on the ordering of input views. 
Instead, we define the world coordinate system with respect to the query view, preserving invariance (to permutations of input views). Finally, we define the scale of the world coordinate system by normalizing all input cameras to have a mean distance of unit length (as in past work~\cite{sargent2024zeronvs}).

\paragraph{Input channel expansion.}
The DiT's input is formed by channel-wise concatenation of three components:
\begin{equation}
    \mathbf{x} = [\mathbf{z}_t;\; \mathbf{y};\; \mathbf{r}] \in \mathbb{R}^{D_{\text{in}} \times (K+1) \times h \times w},
    \label{eq:input}
\end{equation}
where $\mathbf{z}_t$ is the noisy latent, $\mathbf{y}$ is the VAE-encoded embedding concatenated with a binary mask indicating input vs.\ target frames, and $\mathbf{r}$ is the Pl\"ucker ray map.
The patch embedding layer (a 3D convolution) is re-initialized to accommodate the expanded channel count, while all other pretrained weights are preserved.

\paragraph{Zero-Temporal Positional Encoding.}

Standard video diffusion transformers employ Rotary Positional Embeddings (RoPE)~\cite{su2024roformer} to encode temporal order.
While appropriate for video generation, such encodings implicitly assume a sequential structure and introduce ordering bias across frames.
In sparse-view NVS, the input views form an unordered set; injecting temporal position embeddings would therefore violate permutation invariance.

To preserve invariance, we remove temporal positional encoding entirely and treat the $(K+1)$ latent tokens as an unordered set.
As a result, geometric identity is provided solely through camera conditioning via Pl\"ucker ray maps.
This design encourages the transformer to reason over views based on camera geometry rather than frame index.
In practice, we observe that removing temporal positional encoding slightly slows early training convergence, but leads to more stable multi-view generalization.

\paragraph{CLIP and text Embedding.} To preserve permutation invariance, we additionally disable the first-frame CLIP image conditioning used in the original Wan2.1-I2V model, while using an empty text prompt for global text conditioning.

\subsection{Training}
\label{sec:training}

\paragraph{Dataset and sampling.}
We train on multi-view images from a subset of DL3DV-10K~\cite{ling2024dl3dv}, which provides diverse real-world scenes with known camera parameters.
At each iteration, we sample $K{+}1$ frames from a scene: $K$ input views and 1 target view.
We employ a probabilistic sampling strategy: with 80\% probability, frames are sampled uniformly in random order from the full video; with 20\% probability, frames are sampled from a local temporal window to encourage learning from nearby viewpoints.

\paragraph{LoRA fine-tuning.}
Rather than full fine-tuning, we apply Low-Rank Adaptation (LoRA)~\cite{hu2022lora} to the attention and feed-forward layers of the pretrained DiT.
Due to the change in input dimensionality, we re-initialize the patch embedding layer and train it with full gradients.
This strategy preserves the pretrained video generation prior while efficiently adapting the model for camera-conditioned NVS.

\paragraph{Objective.}
We train with the standard flow-matching loss~\cite{lipman2022flow} applied \emph{only} to the target frame:
$
    \mathcal{L} = \mathbb{E}_{t, \boldsymbol{\epsilon}, \mathbf{z}_0} \left\| \boldsymbol{\epsilon}_\theta(\mathbf{z}_t, t, \mathbf{c}) - (\boldsymbol{\epsilon} - \mathbf{z}_0) \right\|^2_{\text{tgt}},
    \label{eq:loss}
$
where $\mathbf{c}$ denotes all conditioning signals (input latents, mask, ray maps, and text embeddings), and $\|\cdot\|^2_{\text{tgt}}$ indicates that the loss is masked to the target frame only.

\section{Experiments}

\subsection{Experiment Setup}

\paragraph{Implementation Details.}
We adopt Wan2.1-I2V-14B~\cite{wan2025} as our video diffusion backbone. 
Training is conducted in two stages: we first train the model at a resolution of $192{\times}336$, and then at a higher resolution of $480{\times}832$ to enhance visual fidelity. 
Low-Rank Adaptation (LoRA) with rank $r{=}32$ is applied to the DiT backbone, and only the patch embedding layer and LoRA parameters are updated, while all other pretrained weights remain frozen. 
This lightweight adaptation strategy ensures that only a small fraction ($\sim 1\%$) of parameters are trained, significantly reducing training cost compared to full fine-tuning baselines. 
We also extend the model to support multi-target prediction via mixed $m$-to-$n$ training 
(see Sec.~\ref{sec:appendix_mton}).

\paragraph{Benchmark and Metrics.}
We evaluate our method on DL3DV-Benchmark~\cite{ling2024dl3dv} and Mip-NeRF 360~\cite{barron2022mip}, which contain diverse real-world scenes with calibrated camera poses and multi-view images. 
For fair comparison, we follow the train/test split from SEVA~\cite{zhou2025stable}. 
We report performance using standard image reconstruction and perceptual metrics, including PSNR, SSIM, LPIPS~\cite{zhang2018unreasonable}, and DreamSim~\cite{fu2023dreamsim}. 
PSNR measures pixel-level fidelity, SSIM structural similarity, while LPIPS and DreamSim quantify perceptual similarity in learned feature spaces.

\paragraph{Baselines.}
We compare against representative generative NVS approaches. 
(1) EscherNet~\cite{kong2024eschernet}, a scalable view synthesis model built upon image diffusion priors, which we fine-tune on the 10K scene-level multi-view dataset (DL3DV~\cite{ling2024dl3dv}) following its original training protocol. 
(2) Aether~\cite{zhu2025aether}, a recent action-conditioned video diffusion model designed for camera-controlled generation. 
(3) SEVA~\cite{zhou2025stable}, state-of-the-art generative method which adapts an image diffusion model to multi-view synthesis using large-scale pose-annotated data.

\subsection{Quantitative Results}
We evaluate all methods under a unified protocol. As SEVA produces square images at $576 \times 576$ and video diffusion models operate at different aspect ratios, we resize or center-crop all predictions to 
$480 \times 480$ for metric computation, ensuring a fair comparison without modifying the original generation settings.

Tab.~\ref{tab:main_comparison} reports quantitative results on the 6-view NVS setting.

\begin{table}[t]
  \caption{
    \textbf{Quantitative comparison on 6-view NVS on DL3DV-Benchmark and Mip-NeRF 360.}
    FrameCrafter uses pretrained video diffusion backbones and is trained on only 1K multi-view scenes.
    Scaling the backbone (CogVideoX-5B → Wan2.1-14B) consistently improves performance, achieving competitive results with prior state-of-the-art methods.
    Best results are highlighted as 
    \colorbox{rankone}{first}, 
    \colorbox{ranktwo}{second}, 
    \colorbox{rankthree}{third}. DS denotes DreamSim~\cite{fu2023dreamsim}. For completeness, we also compare to regression baselines (top two rows) which tend to consistently output blurrier images that produce poor LPIPS and DS but better PSNR and SSIM.
    }
  \label{tab:main_comparison}
  \centering
  \resizebox{\textwidth}{!}{
  \begin{tabular}{lcc|cccc|cccc}
    \toprule
    \multirow{2}{*}{Method} & \multirow{2}{*}{Backbone} & \multirow{2}{*}{\#Scenes}
    & \multicolumn{4}{c|}{DL3DV}
    & \multicolumn{4}{c}{Mip360} \\
    \cmidrule(lr){4-11}
    & & 
    & PSNR $\uparrow$ & SSIM $\uparrow$ & LPIPS $\downarrow$ & DS $\downarrow$
    & PSNR $\uparrow$ & SSIM $\uparrow$ & LPIPS $\downarrow$ & DS $\downarrow$ \\
    \midrule
    LVSM~\cite{jin2024lvsm} & -- & -- & 17.09 & 0.478 & 0.333 & 0.204 & 15.25 & 0.317 & 0.609 & 0.577\\
    E-RayZer~\cite{zhao2025rayzer} & -- & -- & 16.85 & 0.442 & 0.455 & 0.254 & 16.56 & 0.343 & 0.621 & 0.340 \\
    \midrule
    EscherNet~\cite{kong2024eschernet} & SD1.5~\cite{rombach2022high} & 10K & 12.07 & 0.251 & 0.484 & 0.227 & 11.14 & 0.126 & \cellcolor{rankthree} 0.540 & 0.315\\
    Aether~\cite{zhu2025aether} & CogVideoX-5B~\cite{yang2024cogvideox} & -- & 12.66 & 0.258 & 0.469 & 0.140 & 12.60 & \cellcolor{rankthree} 0.220 & 0.651 & 0.334 \\
    SEVA~\cite{zhou2025stable} & SD2.1~\cite{rombach2022high} & 80K & \cellcolor{ranktwo} 16.15 & \cellcolor{rankone}0.470 & \cellcolor{ranktwo} 0.253 & \cellcolor{ranktwo} 0.088 & \cellcolor{ranktwo} 14.59 & \cellcolor{rankone} 0.294 & \cellcolor{ranktwo} 0.372 & \cellcolor{ranktwo} 0.137\\
    Ours & CogVideoX-5B~\cite{yang2024cogvideox} & 1K & \cellcolor{rankthree} 13.65 & \cellcolor{rankthree} 0.272 & \cellcolor{rankthree} 0.389 & \cellcolor{rankthree} 0.113 & \cellcolor{rankthree} 13.09 & 0.209 & \cellcolor{rankthree} 0.540 &  \cellcolor{rankthree}0.204\\
    \textbf{Ours} & Wan2.1-14B~\cite{wan2025} & 1K & \cellcolor{rankone} 17.18 & \cellcolor{ranktwo} 0.445 & \cellcolor{rankone} 0.223 & \cellcolor{rankone} 0.066 & \cellcolor{rankone} 15.64 & \cellcolor{ranktwo} 0.279 & \cellcolor{rankone} 0.365 & \cellcolor{rankone} 0.111\\
    \bottomrule
  \end{tabular}
  \vspace{-104pt}
  }
\end{table}

We first compare with EscherNet~\cite{kong2024eschernet}.
Despite using only 1K training scenes and updating only a small fraction of parameters via LoRA, our model substantially outperforms EscherNet across all metrics. 
These results indicate that video diffusion priors provide significantly stronger multi-view consistency than image-based generative priors under sparse-view settings.

Compared to Aether~\cite{zhu2025aether}, our method achieves consistent improvements when using the same CogVideoX-5B backbone. 
Aether is a camera-conditioned video diffusion model that also incorporates camera ray maps via channel concatenation and generates frames along a specified camera trajectory, for example by conditioning on the first and last frames and interpolating intermediate poses. 
While such a formulation can in principle include the query pose, it assumes a coherent temporal sequence and relies on temporally compressed video encodings, which makes precise pose control more difficult. 
In contrast, our formulation explicitly treats the inputs as an unordered set of views and preserves per-view representations, enabling more accurate geometry-aware generation for sparse-view NVS. 
When scaling to Wan2.1-14B, performance further improves substantially, highlighting the importance of stronger pretrained video backbones.

Most notably, we compare against SEVA~\cite{zhou2025stable}, the state-of-the-art image diffusion-based method, which adapts image diffusion models for multi-view synthesis using substantially larger supervision, including approximately 80K scene-level samples and 340K object-level samples. 
Despite being trained on only 1K scene-level samples -- a large disparity in supervision scale -- our model achieves competitive and often superior performance.
With the Wan2.1-14B backbone, our method surpasses SEVA on PSNR, LPIPS, and DreamSim on DL3DV, while also achieving the best DreamSim and PSNR on Mip-NeRF 360. These results suggest that video diffusion priors provide strong geometric and perceptual cues for sparse-view synthesis, enabling high-quality predictions even with significantly less multi-view supervision.

These results strongly support our central claim: large-scale video diffusion models already encode powerful geometric and cross-view priors, and sparse-view NVS can be achieved through lightweight specialization rather than massive multi-view training. 
Furthermore, the consistent improvement from CogVideoX-5B to Wan2.1-14B reveals a clear backbone scaling trend. As video diffusion models continue to advance, our adaptation framework with lightweight training provides a direct and efficient pathway for transferring these gains to sparse-view novel view synthesis.

We also include two regression-based models as reference. 
Regression-based models often achieve higher PSNR and SSIM due to direct pixel supervision, but typically produce blurrier or less perceptually realistic results in such cases, which is reflected by worse LPIPS and DreamSim scores.
Qualitative comparisons further illustrate this difference (see Sec.~\ref{sec:qualitative}). More quantitative results are provided in the supplementary.

\subsection{Qualitative Results.}
\label{sec:qualitative}
\begin{figure}[t]
    \centering
    \includegraphics[width=\linewidth]{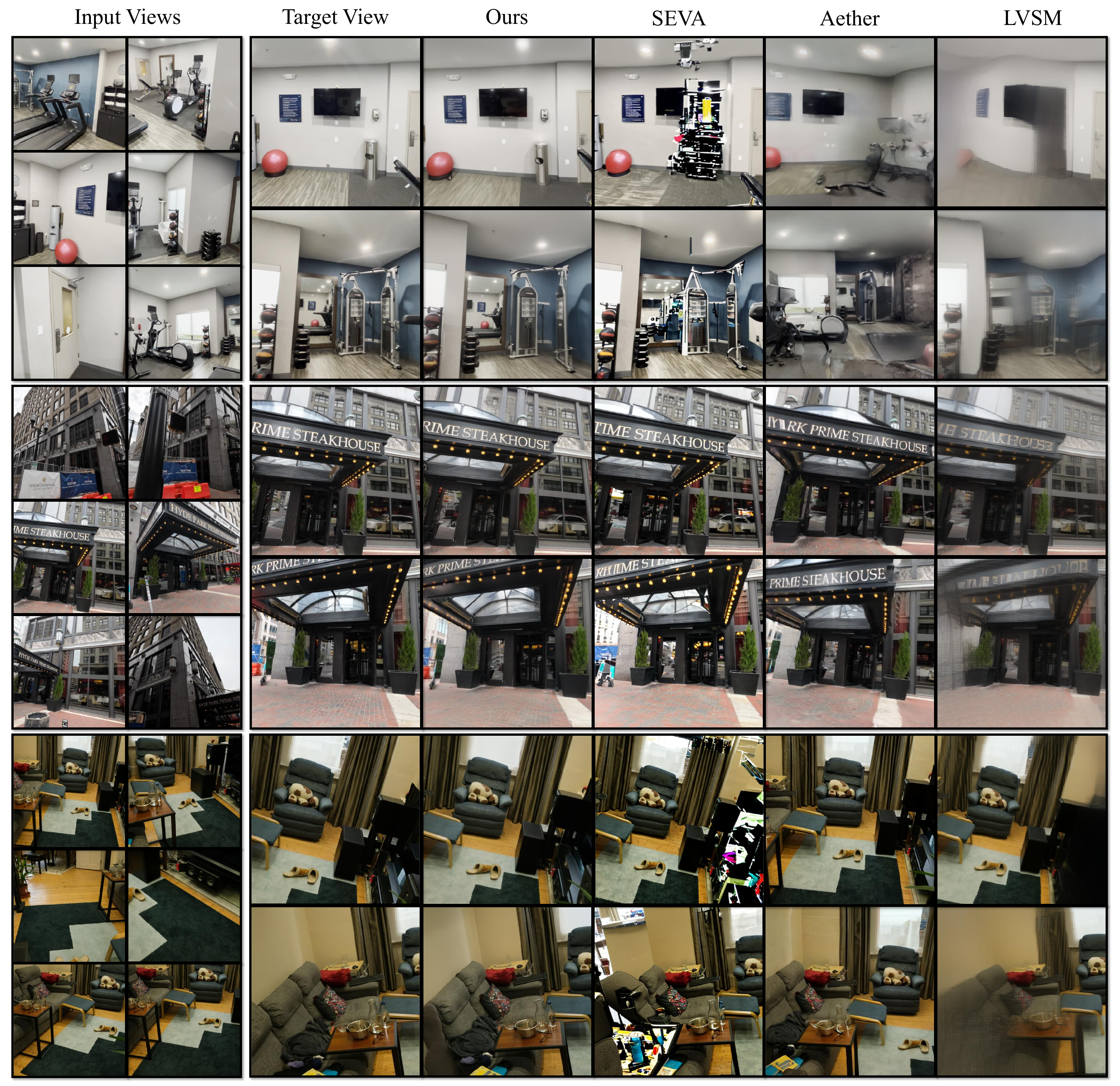}
    \vspace{-20pt}
    \caption{\textbf{Qualitative comparison under sparse inputs.}
    FrameCrafter produces more geometrically consistent and sharper novel views compared to SEVA~\cite{zhou2025stable}, Aether~\cite{zhu2025aether}, LVSM~\cite{jin2024lvsm}, with improved viewpoint alignment and detail preservation.}
    \label{fig:compare1}
    \vspace{-15pt}
\end{figure}

We present qualitative comparisons in Fig.~\ref{fig:compare1}. 
Under sparse-view inputs, our model is able to recover plausible geometry even when certain structures appear only in small regions of the input views, whereas SEVA occasionally fails to maintain geometric consistency. 
Compared with Aether, our method produces more accurate viewpoint alignment and sharper visual details, which we attribute to the per-view independent encoding and our formulation of sparse unordered views within the video diffusion backbone. 
LVSM tends to produce overly smooth predictions with noticeable blur and loss of fine structures, particularly when synthesizing views that require extrapolation from sparse observations. More qualitative comparisons are provided in 
Sec.~\ref{sec:appendix_additional_qualitative}.

\subsection{Ablation Study}
\subsubsection{Design Components}

Tab.~\ref{tab:design_choice} evaluates the contributions of pixel unshuffle, per-view independent encoding, supervision strategy, and positional encoding.

\begin{table}[t]
    \caption{\textbf{Ablation study} on $192 \times 336$ resolution generation (on DL3DV-Benchmark). 
    Removing the per-view independent encoding, pixel unshuffle, or target-only supervision degrades performance, with per-view ``single-frame'' video encodings being the most crucial design choice.}
  \label{tab:design_choice}
  \vspace{-8pt}
  \centering
  \begin{tabular}{@{}l|cccc@{}}
    \toprule
    Variant & PSNR $\uparrow$ & SSIM $\uparrow$ & LPIPS $\downarrow$ & DreamSim $\downarrow$ \\
    \midrule
    \xmark~Per-View Encoding & 11.50 & 0.147 & 0.676 & 0.656 \\
    \xmark~Pixel Unshuffle & 13.53 & 0.212 & 0.359 & 0.129 \\
    \xmark~Target-Only Supervision & 15.08 & 0.287 & 0.270 & 0.116 \\
    \midrule
    w/ PRoPE & 14.91 & 0.289 & 0.315 & 0.174 \\
    w/ Original RoPE & 15.30 & 0.300 & 0.265 & 0.122 \\
    \midrule
    \rowcolor{gray!10}
    \textbf{Full model} & \textbf{15.69} & \textbf{0.326} & \textbf{0.246} & \textbf{0.114} \\
    \bottomrule
  \end{tabular}
  \vspace{-10pt}
\end{table}

\paragraph{Per-View Independent Encoding.}
We also conducted an experiment where, instead of encoding each view independently, we applied the standard spatio-temporal encoding of the causal video VAE. 
Removing individual view encoding and jointly encoding all frames leads to substantial performance degradation. 
This validates our key design choice to eliminate temporal compression and the ordering bias introduced by causal 3D convolutions, which is particularly important for unordered sparse view inputs.

\paragraph{Pixel Unshuffle.}
Replacing pixel unshuffle~\cite{shi2016real} with standard spatial downsampling significantly degrades performance. 
We attribute this to the fact that unshuffling produces latent features with a higher effective channel dimension, providing richer geometric context that better aligns with pose conditioning. 
This confirms that preserving exact per-pixel ray geometry at the latent resolution is critical for accurate view-conditioned generation.

\paragraph{Supervision Strategy.}
We compare target-only supervision with full-sequence supervision, where input views are also supervised, and find that supervising input views hurts performance.
This is likely because the loss is dominated by reconstructing input frames, reducing emphasis on query view prediction.
This insight may benefit other conditional generation tasks with clean input frames~\cite{van2026anyview,xiao2025video4spatial}.

\paragraph{Positional Encoding.}
We compare against retaining full spatio-temporal Rotary Positional Encoding (3D RoPE).
Removing the temporal component yields improved performance quality while preserving permutation invariance (see Sec.~\ref{sec:ablation_permutation}).
We attribute this to the fact that disabling temporal RoPE encourages the transformer to rely more on explicit camera conditioning rather than implicit frame index cues, which are unnecessary for unordered sparse-view inputs.
We also experimented with replacing the original RoPE with Projective Positional Encoding (PRoPE)~\cite{li2025cameras}, a relative positional encoding technique that captures complete camera frustums (both intrinsics and extrinsics) in attention conditioning. 
While PRoPE has been shown to improve performance in feedforward NVS models, we did not observe consistent gains in our setting under minimal training (LoRA with limited multi-view data). 
We conjecture that PRoPE modifies the attention structure more significantly, and fully adapting the pretrained backbone to this change may require larger-scale retraining.

Overall, each component contributes meaningfully, and their combination achieves the best reconstruction and perceptual performance.

\subsubsection{Permutation Invariance}
\label{sec:ablation_permutation}
We evaluate the sensitivity of different design choices to input ordering.
Specifically, we compare: (i) a first-view coordinate formulation (where poses are normalized relative to the first input view), 
(ii) model using standard spatio-temporal RoPE, and (iii) our design with temporal RoPE removed and target-view coordinate frame.
At test time, we evaluate each model under 10 random permutations of the same input set on DL3DV-Benchmark~\cite{ling2024dl3dv}. We report the mean performance across all runs together with the mean standard deviation across permutations (Tab.~\ref{tab:permute}). 

\begin{table}[h]
\vspace{-15pt}
\caption{\textbf{Performance under 10 random permutations of input views.}}
\label{tab:permute}
\vspace{-10pt}
\centering
\resizebox{\textwidth}{!}{
\begin{tabular}{lcc|cc|cc|cc}
\toprule
\multirow{2}{*}{Training Setup} 
& \multicolumn{2}{c|}{PSNR} 
& \multicolumn{2}{c|}{SSIM} 
& \multicolumn{2}{c|}{LPIPS} 
& \multicolumn{2}{c}{DreamSim} \\
\cmidrule(lr){2-3} \cmidrule(lr){4-5} \cmidrule(lr){6-7} \cmidrule(lr){8-9}
& Mean $\uparrow$ & Std $\downarrow$ & Mean $\uparrow$ & Std $\downarrow$ & Mean $\downarrow$ & Std $\downarrow$ & Mean $\downarrow$ & Std $\downarrow$ \\
\midrule
First-view Coordinate & 15.67 & 0.4730 & 0.309 & 0.0276 & 0.262 & 0.0203 & 0.112 & 0.0111\\
w/ original RoPE  & 15.61 & 0.3106 & 0.310 & 0.0212 & 0.254 & 0.0095 & 0.115 & 0.0063 \\
\textbf{Ours} & \textbf{16.21} & \textbf{0.0081} & \textbf{0.348} & \textbf{0.0006} & \textbf{0.224} & \textbf{0.0003} & \textbf{0.102} & \textbf{0.0005} \\
\bottomrule
\end{tabular}
}
\vspace{-20pt}
\end{table}

Models that rely on a first-view coordinate exhibit degraded performance and high variance, suggesting that this design introduces bias that harms generalization.
Using standard temporal RoPE yields similar performance, but still shows non-negligible variance across permutations due to the implicit ordering bias in temporal positional encoding. 
Removing the temporal component of RoPE and adopting a target-view coordinate frame architecturally eliminate input-order bias, further improving mean performance and significantly reducing permutation variance, demonstrating strong robustness to input reordering.

One may ask why permutation invariance is important for sparse-view NVS, and whether it is possible to simply choose a suitable ordering of the input views. To further investigate this question, we report the performance when the model is trained and evaluated using an ``ordered'' input sequence (w/ temp RoPE).


As shown in Tab.~\ref{tab:permutation_setting}, even when the model is trained and evaluated using ordered inputs, its performance remains worse than the model trained with randomly shuffled inputs under the same ordered evaluation. This compares two possible problem formulations: if sparse-view NVS is treated as an ordered sequence modeling problem, the ordered model represents a natural upper attempt under this setting; however, if the inputs are instead treated as an unordered set, random permutations become valid training samples. The improvement suggests that sparse-view observations do not naturally correspond to a smooth temporal trajectory: input cameras often exhibit large viewpoint changes and irregular spatial distributions around the scene, making it difficult to define a meaningful global ordering. 
Therefore, permutation-invariant architecture and training better match the nature of sparse-view NVS. Also this formulation can serves as an effective form of data augmentation, since each scene can be observed under many equivalent input orderings, exposing the model to a much richer set of training configurations. Permutation-invariant training can be viewed as implicitly leveraging a factorially larger set of permuted training samples.


\vspace{-10pt}
\begin{table}[h]
  \caption{\textbf{Effect of training setup under ordered evaluation} on $192 \times 336$ resolution 6-view NVS (DL3DV-Benchmark~\cite{ling2024dl3dv}). 
    During evaluation, input views follow their original temporal order in the source video.}
  \label{tab:permutation_setting}
  \vspace{-8pt}
  \centering
  \begin{tabular}{@{}lcccccc@{}}
    \toprule
    Training Setting & PSNR $\uparrow$ & SSIM $\uparrow$ & LPIPS $\downarrow$ & DreamSim $\downarrow$ \\
    \midrule
    Ordered & 15.04 & 0.292 & 0.276 & 0.136\\
    \textbf{Ours (Permutation-Invariant)} & \textbf{15.69} & \textbf{0.326} & \textbf{0.246} & \textbf{0.114} \\
    \bottomrule
  \end{tabular}
  \vspace{-10pt}
\end{table}

Our architectural design naturally supports this behavior. In particular, removing the temporal component of RoPE, using per-view independent VAE encoding, and adopting query-centered camera coordinate normalization together ensure that the model does not rely on a specific ordering of the input views. This allows the network to reason more effectively over unordered sets of views and improves robustness in sparse-view settings.

\subsubsection{Data Scaling}
\begin{wrapfigure}{r}{0.5\textwidth}
    \vspace{-35pt}
    \includegraphics[width=\linewidth]{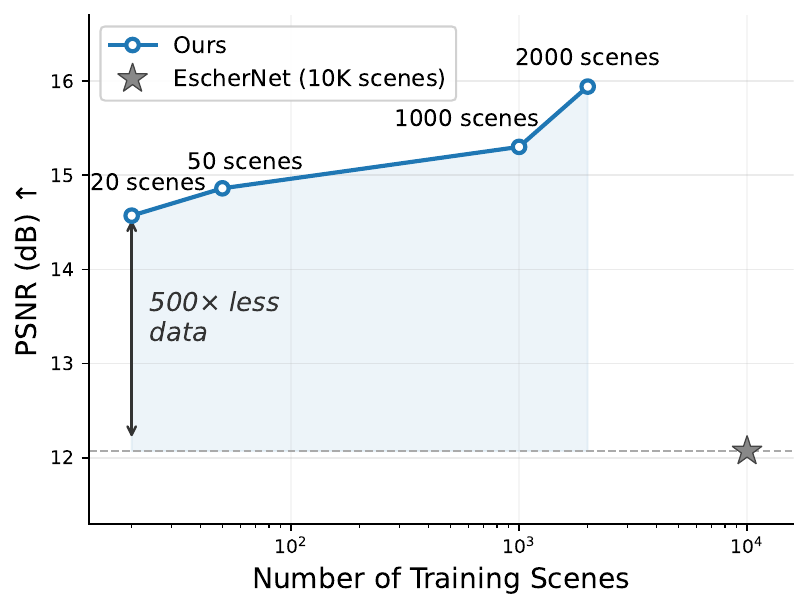}
    \vspace{-20pt}
    \caption{\textbf{Performance on different numbers of training scenes.}
    Even with only 20 scenes, our model surpasses EscherNet trained on 10K scenes.
    Performance improves consistently as training data increases, demonstrating strong data efficiency and scalability.
    }
    \label{fig:ablation_data}
    \vspace{-20pt}
\end{wrapfigure}
We further analyze the effect of training data scale on model performance (Fig.~\ref{fig:ablation_data}) to examine the limits of minimal-data adaptation. 
Remarkably, even when trained on only 20 scenes, our model surpasses EscherNet trained on 10K scenes. 
This highlights the strength of video diffusion models as geometric priors: the lightweight training primarily adapts the model to the sparse-view setting and camera conditioning, whereas multi-view reasoning appears to already emerge from large-scale video pretraining.

As the number of training scenes increases, performance improves consistently, revealing a clear scaling trend.
These results suggest that video diffusion priors provide a strong geometric foundation, enabling sparse-view NVS to benefit from additional data without requiring large-scale multi-view supervision.




\section{Conclusion}
We revisit sparse-view novel view synthesis from the perspective of video generation. 
Instead of adapting image diffusion models with large-scale multi-view supervision, we demonstrate that modern video diffusion models already encode strong geometric and multi-view priors. 
By introducing per-view independent encoding, query-centered Pl\"ucker-based geometric conditioning, temporal RoPE removal and lightweight LoRA adaptation, we transform a pretrained video diffusion backbone into an effective permutation-invariant sparse-view NVS model using only 1K training scenes.

Our results show that this approach achieves competitive performance against state-of-the-art methods trained with substantially larger multi-view datasets. Moreover, performance improves consistently as the video backbone scales, suggesting that advances in large-scale video diffusion models can be directly transferred to novel view synthesis without requiring massive curated multi-view data. These findings point toward a scalable paradigm for NVS: rather than treating multi-view synthesis as a standalone modeling problem, it can be viewed as a lightweight specialization of increasingly powerful video foundation models.

Our analysis also suggests a provocative take on the knowledge emergent in video generative ``world models.'' Indeed, we were surprised by the ease with which such models can be tuned to be permutation-invariant, losing any sense of temporal dynamics. This suggests that current models might capture more ``multi-view'' knowledge about the world, rather than the dynamics of how it evolves over time. At the very least, such temporal knowledge is surprisingly easy to unlearn. We hope our work leads to more exploration of such representational questions. More broadly, it points to the potential of adapting pretrained video models to non-temporal tasks and using video diffusion priors as alternatives to image diffusion priors.

\vspace{10pt}
\small \noindent \textbf{Acknowledgments:} 
We thank Prof. Shubham Tulsiani for insightful discussions and perspectives, and Nikhil Keetha and other members of Deva's lab at CMU for valuable feedback and suggestions.

This work used Bridges-2 at Pittsburgh Supercomputing Center through allocation \texttt{cis250177p} from the Advanced Cyberinfrastructure Coordination Ecosystem: Services \& Support (ACCESS) program, which is supported by National Science Foundation grants \#2138259, \#2138286, \#2138307, \#2137603, and \#2138296.
This work was supported by Intelligence Advanced Research Projects Activity (IARPA) via Department of Interior/Interior Business Center (DOI/IBC) contract number 140D0423C0074. The U.S. Government is authorized to reproduce and distribute reprints for Governmental purposes notwithstanding any copyright annotation thereon. Disclaimer: The views and conclusions contained herein are those of the authors and should not be interpreted as necessarily representing the official policies or endorsements, either expressed or implied, of IARPA, DOI/IBC, or the U.S. Government.



%
%
\bibliographystyle{splncs04}
\bibliography{main}
\appendix

\begin{center}
{\Large \textbf{Appendix}}
\end{center}

\noindent\textbf{Overview.}
In the appendix, we provide additional quantitative/qualitative results, ablation studies, and implementation details:
\begin{itemize}
    \item \textbf{Sec.~\ref{sec:appendix_additional_quantitative}}: Additional quantitative results, including analyses under different numbers of input and output views, comparison with more video diffusion baselines, compute time, and downstream 3D reconstruction consistency.
    \item \textbf{Sec.~\ref{sec:appendix_additional_qualitative}}: Additional qualitative results.
    \item \textbf{Sec.~\ref{sec:appendix_implementation}}: More implementation details.
\end{itemize}

\section{Additional Quantitative Results}
\label{sec:appendix_additional_quantitative}

\subsection{Generalization to Different Numbers of Input Views}

We also evaluate our model under a more challenging 3-view setting. 
The results are reported in Tab.~\ref{tab:add_comparison}. 
Despite not being specifically trained for this scenario, our model generalizes well to this setting and still achieves strong performance, demonstrating robustness to different numbers of input views.

\begin{table}[h]
  \caption{
    \textbf{Quantitative comparison on 3-view NVS on DL3DV-Benchmark~\cite{ling2024dl3dv} and Mip-NeRF 360~\cite{barron2022mip}.}
    Best results are highlighted as 
    \colorbox{rankone}{first}, 
    \colorbox{ranktwo}{second}, 
    \colorbox{rankthree}{third}. DS denotes DreamSim~\cite{fu2023dreamsim}. All outputs are resized or center-cropped to $480\times 480$ for fair comparisons.
    }
  \label{tab:add_comparison}
  \centering
  \resizebox{\textwidth}{!}{
  \begin{tabular}{lcc|cccc|cccc}
    \toprule
    \multirow{2}{*}{Method} & \multirow{2}{*}{Backbone} & \multirow{2}{*}{\#Scenes}
    & \multicolumn{4}{c|}{DL3DV}
    & \multicolumn{4}{c}{Mip360} \\
    \cmidrule(lr){4-11}
    & & 
    & PSNR $\uparrow$ & SSIM $\uparrow$ & LPIPS $\downarrow$ & DS $\downarrow$
    & PSNR $\uparrow$ & SSIM $\uparrow$ & LPIPS $\downarrow$ & DS $\downarrow$ \\
    \midrule
    EscherNet~\cite{kong2024eschernet} & SD1.5~\cite{rombach2022high} & 10K & 11.43 & 0.228 & \cellcolor{rankthree} 0.524 & 0.266 & 10.44 & 0.171 & 0.627 & 0.383\\
    Aether~\cite{zhu2025aether} & CogVideoX-5B~\cite{yang2024cogvideox} & -- & \cellcolor{rankthree} 11.98 & \cellcolor{rankthree} 0.245 & 0.531 & \cellcolor{rankthree} 0.204 & \cellcolor{ranktwo} 13.11 & \cellcolor{rankthree} 0.228 & \cellcolor{rankthree} 0.614 & \cellcolor{rankthree} 0.239 \\
    SEVA~\cite{zhou2025stable} & SD2.1~\cite{rombach2022high} & 80K & \cellcolor{ranktwo} 13.52 & \cellcolor{rankone}0.359 & \cellcolor{ranktwo} 0.364 & \cellcolor{ranktwo} 0.167 & \cellcolor{rankthree} 12.73 & \cellcolor{rankone} 0.242 & \cellcolor{rankone} 0.443 & \cellcolor{ranktwo} 0.198\\
    \textbf{Ours} & Wan2.1-14B~\cite{wan2025} & 1K & \cellcolor{rankone} 14.73 & \cellcolor{ranktwo} 0.334 & \cellcolor{rankone} 0.351 & \cellcolor{rankone} 0.129 & \cellcolor{rankone} 14.14 & \cellcolor{ranktwo} 0.235 & \cellcolor{ranktwo} 0.491 & \cellcolor{rankone} 0.166\\
    \bottomrule
  \end{tabular}
  }
\end{table}

\subsection{Flexible Multi-Target Prediction}
\label{sec:appendix_mton}

Our main model is trained for single-target prediction, i.e., generating one novel view given $K=6$ input views. 
When the model is directly applied to generate multiple target views simultaneously, performance gradually degrades as the number of predicted views increases (Tab.~\ref{tab:appendix_mton}). 
This behavior is expected, since the model is optimized only for single-target prediction and therefore not trained to jointly reason over multiple query views.

To enable flexible multi-target inference, we further train the model using a mixed input–output formulation with a fixed temporal length $T=10$. 
For each training sample, the number of input views is randomly sampled between 3 and 9, and the remaining frames are treated as target views. 
This setup exposes the model to a wide range of input–output configurations and enables it to support general $m$-to-$n$ novel view synthesis.

After this additional training, the performance gap across different 6-to-$n$ settings largely disappears (Tab.~\ref{tab:appendix_mton}). 
The model achieves similar quality when generating multiple target views simultaneously, and the training also benefits the single-target setting.

We attribute this improvement to the additional geometric constraints introduced when predicting multiple target views simultaneously. 
Since these views correspond to different camera poses of the same scene, jointly modeling them encourages the network to learn representations that remain consistent under viewpoint changes. 
This effectively provides stronger multi-view supervision during training, acting as a form of regularization that improves the model's ability to reason about scene geometry and view-dependent appearance. 

\begin{table}[t]
\caption{\textbf{Flexible multi-target prediction.}
Performance comparison before and after training with mixed $m$-to-$n$ supervision ($T=10$). 
Evaluation is performed at $192 \times 336$ resolution on DL3DV-Benchmark~\cite{ling2024dl3dv}.}
\label{tab:appendix_mton}
\vspace{-8pt}
\centering
\resizebox{\linewidth}{!}{
\begin{tabular}{lcccc|cccc}
\toprule
\multirow{2}{*}{Setting} 
& \multicolumn{4}{c|}{Original} 
& \multicolumn{4}{c}{After Training} \\
\cmidrule(lr){2-5} \cmidrule(lr){6-9}
& PSNR $\uparrow$ & SSIM $\uparrow$ & LPIPS $\downarrow$ & DreamSim $\downarrow$
& PSNR $\uparrow$ & SSIM $\uparrow$ & LPIPS $\downarrow$ & DreamSim $\downarrow$ \\
\midrule
6-to-1 & 15.69 & 0.326 & 0.246 & 0.114 & 16.27 & 0.365 & 0.226 & 0.100 \\
6-to-2 & 14.84 & 0.261 & 0.347 & 0.200 & 16.20 & 0.374 & 0.229 & 0.102 \\
6-to-3 & 14.62 & 0.239 & 0.419 & 0.280 & 16.42 & 0.374 & 0.223 & 0.096 \\
6-to-4 & 14.18 & 0.217 & 0.511 & 0.384 & 16.23 & 0.367 & 0.225 & 0.097 \\
\bottomrule
\end{tabular}}
\vspace{-10pt}
\end{table}

\subsection{More baselines and compute time}
Most existing video-diffusion baselines for NVS target trajectory-based generation with predefined temporal camera paths, often relying on explicit 3D reconstruction from off-the-shelf models. In contrast, our setting is set-based $N$-to-$1$/$N$-to-$M$ sparse-view NVS, where the input views are unordered and permutation-invariant. To our knowledge, no prior work applies video diffusion to permutation-invariant NVS in this setting. For completeness, we adapt these trajectory-based methods to our benchmark using a two-closest-view interpolation setup, and report both reconstruction quality and inference time in Tab.~\ref{tab:timesteps_cost}. Although our method is slower than SEVA at larger sampling budgets, it substantially reduces training cost by leveraging pretrained video diffusion priors: we use LoRA fine-tuning for only 30K iterations on 1K scenes, with 7.8$\times$ fewer trainable parameters, 30$\times$ fewer iterations, and 80$\times$ less data than SEVA. Our method is also much faster than trajectory-based video-diffusion baselines and already performs strongly with only 5--10 diffusion steps (fewer steps lead to blurrier results, which favor PSNR and SSIM), with further speedups possible via DMD.

\begin{table}[h]
\centering
\caption{\textbf{Compute time and comparison with baselines} on DL3DV (6-view NVS) using a single A6000 GPU.}
\label{tab:timesteps_cost}
\begin{tabular}{lccccc}
\toprule
Method & PSNR $\uparrow$ & SSIM $\uparrow$ & LPIPS $\downarrow$ & DreamSim $\downarrow$ & Avg. Time $\downarrow$ \\
\midrule
ViewCrafter~\cite{yu2024viewcrafter} & 13.55 & 0.312 & 0.443 & 0.181 & $\sim$247s \\
GEN3C~\cite{ren2025gen3c} + VGGT~\cite{wang2025vggt} & 15.95 & 0.398 & 0.333 & 0.099 &$\sim$1460s \\
SEVA~\cite{zhou2025stable}        & 16.15 & 0.470 & 0.253 & 0.088 & $\sim$24s \\
\midrule
Ours (5 steps)  & 18.13 & 0.485 & 0.298 & 0.114 & $\sim$22s \\
Ours (10 steps) & 17.72 & 0.469 & 0.254 & 0.089 & $\sim$42s \\
Ours (25 steps) & 17.35 & 0.454 & 0.226 & 0.070 & $\sim$99s \\
Ours (50 steps) & 17.18 & 0.445 & 0.223 & 0.066 & $\sim$177s \\
\bottomrule
\end{tabular}
\end{table}

\subsection{3D reconstruction consistency}
We further evaluate downstream 3D consistency on DL3DV under the 6-view NVS setting (Tab.~\ref{tab:3d-recon-consistency}). For each synthesized target frame, we run MapAnything~\cite{keetha2026mapanything} using the input views and the generated target view. We then compare the reconstructed target-view depth/geometry with the reconstruction obtained using the same input views and the GT target image. We report depth metrics (AbsRel, RMSE, and $\delta\!<\!1.25$) and pointmap metrics (accuracy, completeness, and overall (Chamfer-L1)), all computed in metric scale in the target-camera frame.

\begin{table}[h]
\centering
\caption{\textbf{Downstream 3D reconstruction consistency.}
Using the same reconstruction pipeline, our generated views yield lower depth errors and better pointmap accuracy, completeness, and overall error than SEVA.}
\label{tab:3d-recon-consistency}
\setlength{\tabcolsep}{4pt}
\resizebox{\columnwidth}{!}{
\begin{tabular}{l c c c c c c}
\toprule
& \multicolumn{3}{c}{Depth} & \multicolumn{3}{c}{Pointmap} \\
\cmidrule(lr){2-4}\cmidrule(lr){5-7}
Method & AbsRel\,$\downarrow$ & $\delta\!<\!1.25$\,$\uparrow$ & RMSE\,(m)\,$\downarrow$ & Acc.\,(m)\,$\downarrow$ & Comp.\,(m)\,$\downarrow$ & Overall\,(m)\,$\downarrow$ \\
\midrule
SEVA  & 0.110          & \textbf{0.947} & 0.857          & 0.254          & 0.249          & 0.251          \\
Ours  & \textbf{0.073} & 0.940          & \textbf{0.789} & \textbf{0.163} & \textbf{0.188} & \textbf{0.176} \\
\bottomrule
\end{tabular}
}
\end{table}

\section{Additional Qualitative Results}
\label{sec:appendix_additional_qualitative}
\begin{figure}[t]
    \centering
    \includegraphics[width=\linewidth]{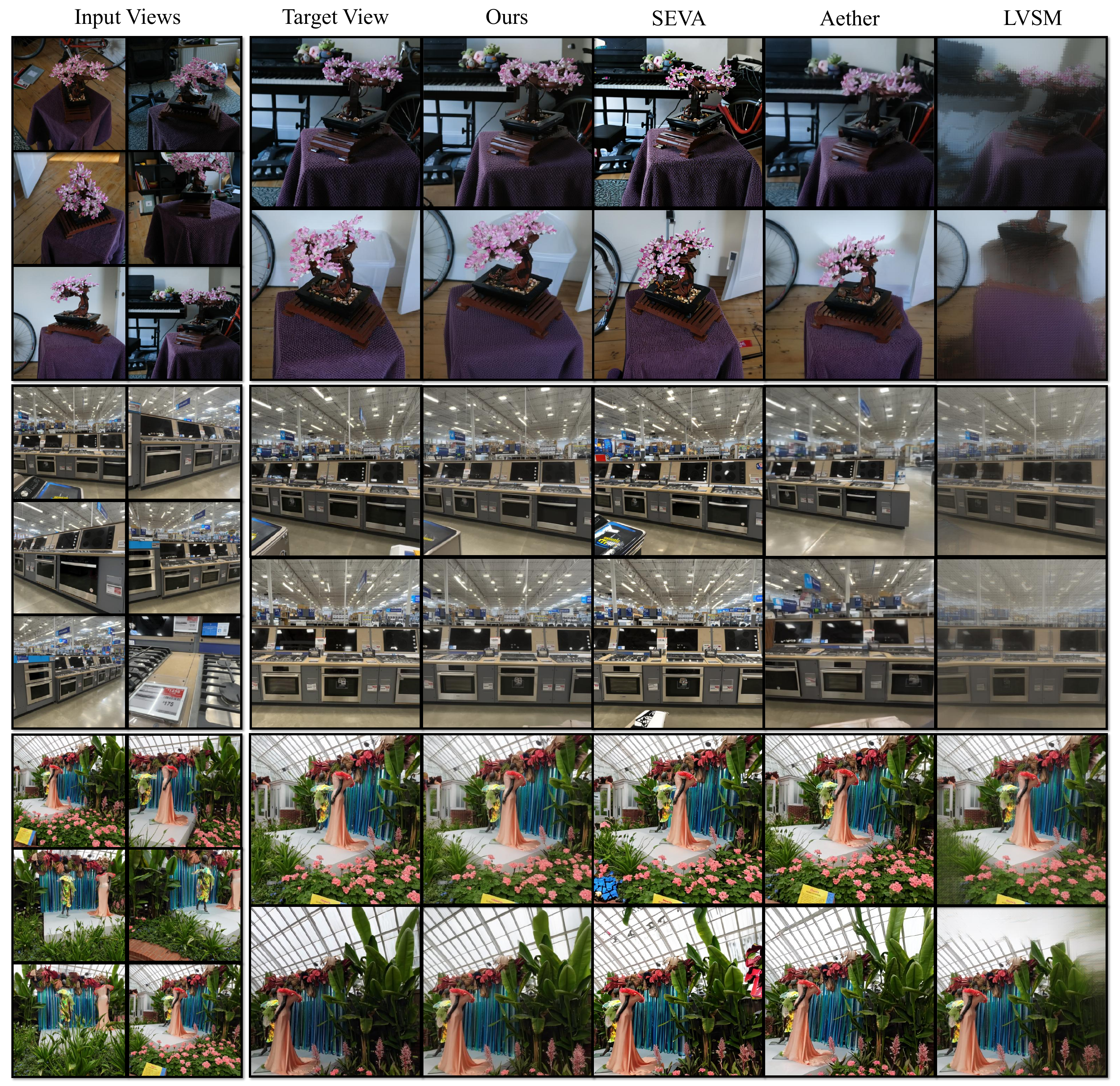}
    \caption{\textbf{Additional Qualitative Comparisons} with SEVA~\cite{zhou2025stable}, Aether~\cite{zhu2025aether} and LVSM~\cite{jin2024lvsm}.}
    \label{fig:compare2}
\end{figure}

We provide additional qualitative comparisons in Fig.~\ref{fig:compare2}. 

\begin{figure}[t]
\centering
\begin{tabular}{cc}
\includegraphics[width=0.48\linewidth]{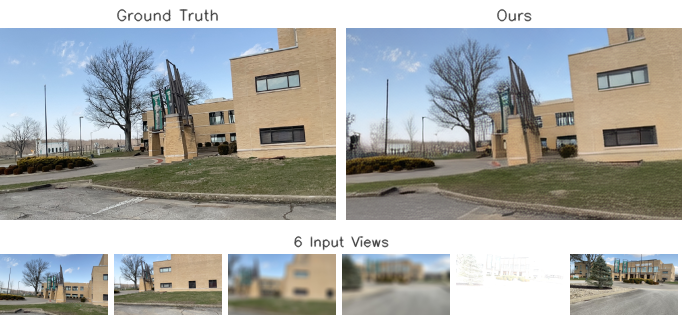} &
\includegraphics[width=0.48\linewidth]{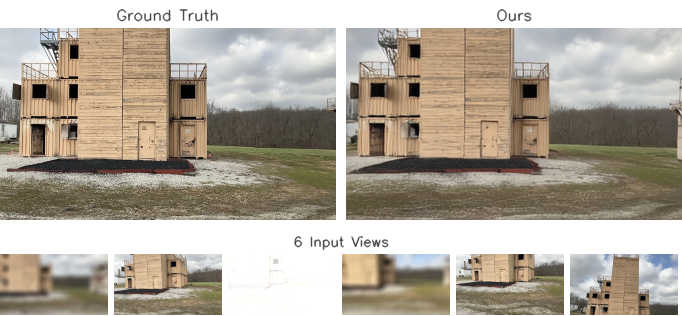} \\
\includegraphics[width=0.48\linewidth]{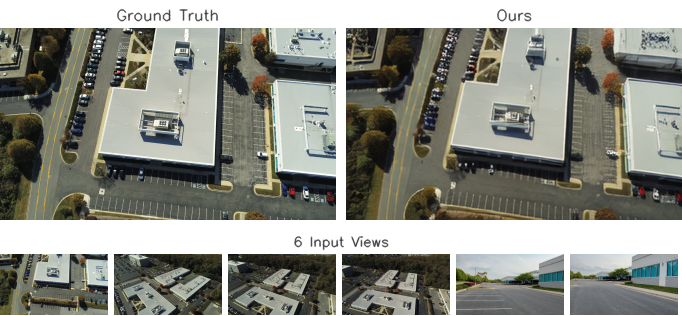} &
\includegraphics[width=0.48\linewidth]{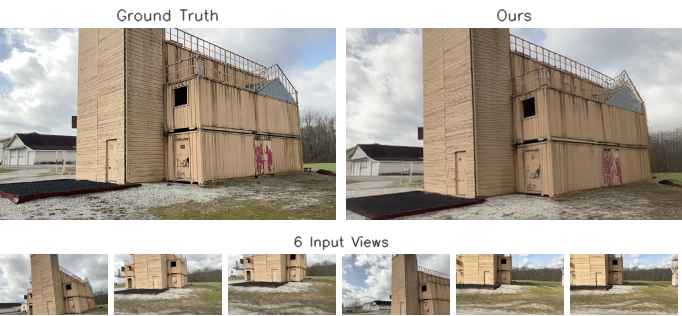}
\end{tabular}
\caption{
\textbf{In-the-wild qualitative results.}
We evaluate FrameCrafter using COLMAP-estimated camera poses, which contain realistic noise and reconstruction artifacts.
The results suggest robustness beyond curated benchmark settings.
}
\label{fig:in-the-wild}
\end{figure}

We further show in-the-wild examples in Fig.~\ref{fig:in-the-wild}, where camera poses are estimated by COLMAP~\cite{schonberger2016structure} and may contain realistic noise and reconstruction artifacts.
These results demonstrate that FrameCrafter remains robust beyond curated benchmarks.

\section{More Implementation Details}
\label{sec:appendix_implementation}
Low-resolution training ($192 \times 336$) is performed using NVIDIA A6000 GPUs, and the high-resolution stage ($480 \times 832$) is trained on NVIDIA H100 GPUs. 
We find that this resolution curriculum—first training at low resolution and then fine-tuning at higher resolution—not only improves training efficiency but also leads to more stable optimization and better final performance.

\end{document}